\documentclass{IEEEtran}

\usepackage{booktabs} 
\usepackage{enumitem}
\usepackage{graphicx}
\usepackage{mathtools}
\usepackage{hyperref}
\usepackage{algorithmic}
\usepackage[ruled,lined,linesnumbered]{algorithm2e} 

\usepackage{mathtools}
\usepackage{cite}

\newtheorem{definition}{Definition}
\newtheorem{example}{Example}

\begin{document}
\title{Learning Unknown Examples For ML Model Generalization
\thanks{}
}

 \author{
 \IEEEauthorblockN{Yeounoh Chung\textsuperscript{1}}
 \quad
 \and
 \IEEEauthorblockN{Peter J. Haas\textsuperscript{2}}
 \quad
 \and
 \IEEEauthorblockN{Tim Kraska\textsuperscript{3}}
 \quad
 \and
 \IEEEauthorblockN{Eli Upfal\textsuperscript{1}}
 \quad
 \\
 \IEEEauthorblockA{\textsuperscript{1}\{first\_last\}@brown.edu}
 \quad
 \IEEEauthorblockA{\textsuperscript{2}phaas@cs.umass.edu}
\quad
\IEEEauthorblockA{\textsuperscript{3}kraska@mit.edu}
 }


%
%
%
%

\maketitle
\begin{abstract}

Most machine learning (ML) technology assumes that the data for training an ML model has the same distribution as the test data to which the model will be applied. However, due to sample selection bias or, more generally, covariate shift, there exist potential training examples that are unknown to the modeler---``unknown unknowns''. The resulting discrepancy between training and testing distributions leads to poor generalization performance of the ML model and hence biased predictions. Existing techniques use test data to detect and ameliorate such discrepancies, but in many real-world situations such test data is unavailable at training time. We exploit the fact that training data often comes from multiple overlapping sources, and combine species-estimation techniques with data-driven methods for estimating the feature values for the unknown unknowns. This information can then be used to correct the training set, prior to seeing any test data. Experiments on a variety of ML models and datasets indicate that our novel techniques can improve generalization performance and increase ML model robustness.

\end{abstract}
\begin{IEEEkeywords}
unknown unknowns, generalization, sampling bias, covariate shift.
\end{IEEEkeywords}
%
%




\section{Introduction}

Over the past decades, researchers and Machine Learning (ML) practitioners have come up with better and better ways to build, understand and improve the quality of ML models, but mostly under the key assumption that the training data is distributed identically to the testing data. This assumption usually holds true in algorithm-development environments and data-science competitions, where a single dataset is split into training and testing sets, but does it hold more generally? If not, what are the consequences for ML?

Experience shows that the foregoing assumption can fail dramatically in many real-world scenarios, especially when the data needs to be collected and integrated over multiple sources and over a long period of time. This issue is well known, for example, to the Census Bureau. A 2016 Census Advisory Committee report~\cite{Census16} highlights the difficulties in reaching groups such as racial and ethnic minorities, poor English speakers, low income and homeless persons, undocumented immigrants, children, and more. Some of these groups do not have access to smartphones or the internet, or they fear interactions with authorities, so the the prospects for data collection will remain difficult into the foreseeable future. Similarly, a recent report on fairness in precision medicine~\cite{FerrymanP18} documents bias in labeled medical datasets and asserts that ``insofar as we still have a systematically describable group who are not in a health care system with data being collected upon them, from them, then that will be a source of bias.'' In each of these cases, factors such as income and ethnicity can result in exclusion of items from a training set, yielding unrepresentative training data. We emphasize that the issue here is not just \emph{underrepresentation} of classes of data items, but the \emph{complete absence} of these items from consideration because they are unknown to the ML modeler.

Besides sampling bias, population shifts over time can lead to unrepresentative training data. For example, a regression model for predicting height based on weight that was trained on the US population in the 80's may not be usable today, because the variables (population-wide height and weight distributions) have changed over time, so that the old training data do not represent the actual testing data of today.

In either case, the unrepresentativeness of the training data will adversely affect an ML model's ability to handle unseen test data. Indeed, the better the fit to biased training data, the harder is becomes for the model to handle new test data. Clearly, mitigation of biased training data is crucial for achieving fair ML.

In this work, we focus on the impact of unknown training instances on ML model performance. The unknown examples during training can arise if the training distribution $p'(x)$ is different from the testing distribution $p(x)$ due to \emph{sample selection bias}---where some data items from the testing distribution are more or less likely to be sampled in the training data---and, more generally, \emph{covariate shift}, where the training data and testing data distributions can be different for any reason.
Although $p(x) \neq p'(x)$, we assume that $p(y|x) = p'(y|x)$, so that the conditional distribution of the class variable $y$ of interest is the same for both training and test data. That is, the predictive relationship between $x$ and $y$ is the same; only the data distribution of the $x$-values differs.

\emph{Generalization} is the ability of a trained ML model to accurately predict on examples that were not used for training \cite{Bishop:2006:PRM:1162264}. Good generalization performance is a key goal of any practical learning algorithm. Ideally, we want to fit the model on a training set that well represents the hidden testing data or the target population---i.e., \emph{the training and testing data are drawn from the same distribution}. If this is not the case, then we end up with ``malignant'' unknown instances that are missing during the training. If the training data is biased, in a sense that some parts of the population are under-represented or missing in the training data (i.e., we have unknown instances), then the fitted model $f$ on that training data will be biased away from the optimal function $f^*$ and have poor generalization performance. 
It is important to note that, even if the training and testing distributions were the same ($p(x) = p'(x)$), we would still have missing examples in the training set due to luck of the draw. We call such missing examples ``benign'' unknown instances because their absence does not cause $p(x)$ to systematically shift away from $p'(x)$ and thus they do not have much harmful impact on model quality. Benign unknown instances can be combatted via additional training data; this strategy fails for malignant instances.

The issue of malignant unknown instances is orthogonal to the typical model complexity/generalization trade-off, where over-fitting or under-fitting the model to the training data---even if it distributed exactly according to $p(x)$---can result in poor generalization. Many techniques, such as cross-validation and regularization~\cite{wiki:overfitting}, have been developed to address this problem. However, even a well-trained model with no under/over-fitting can fail to generalize to the testing set in the presence of covariate shift; 
in Section~\ref{sec:experiment}, we show that both simple and complex models can suffer in the presence of malignant unknown instances.

\subsection{Learning under covariate shift}
\label{sec:intro_covariate_shift}

Learning under covariate shift has been studied extensively \cite{bickel2007discriminative,sugiyama2013learning,zadrozny2004learning,liu2014robust,huang2007correcting}. An important observation from importance sampling states that the accuracy loss on the test distribution can be minimized by weighting the loss on the training distribution with the scaling factor, $p(x)/p'(x)$ \cite{shimodaira2000improving}. The previous work proposes many techniques to estimate the scaling factor or the training, testing or the conditional densities more accurately and efficiently, which in turn, require both training and ``unlabeled'' testing data \cite{huang2007correcting}.

Access to the unlabeled testing data (during training) is only feasible in a setting where the actual test data is provided, e.g., in a data science competition. However, using such a target dataset (or re-training the model after seeing the test data) may not be possible in many real applications. We therefore propose the first techniques for learning under covariate shift that require just the training data, along with the data redundancy information that is typically available (prior to cleaning) when a dataset is integrated from multiple data sources with overlapping information. As discussed later on, the number of data items with low redundancy contains important information about the number of unknown data items that were not included in the training data.


\begin{figure}[t]
 \centering
 \includegraphics[width=\columnwidth]{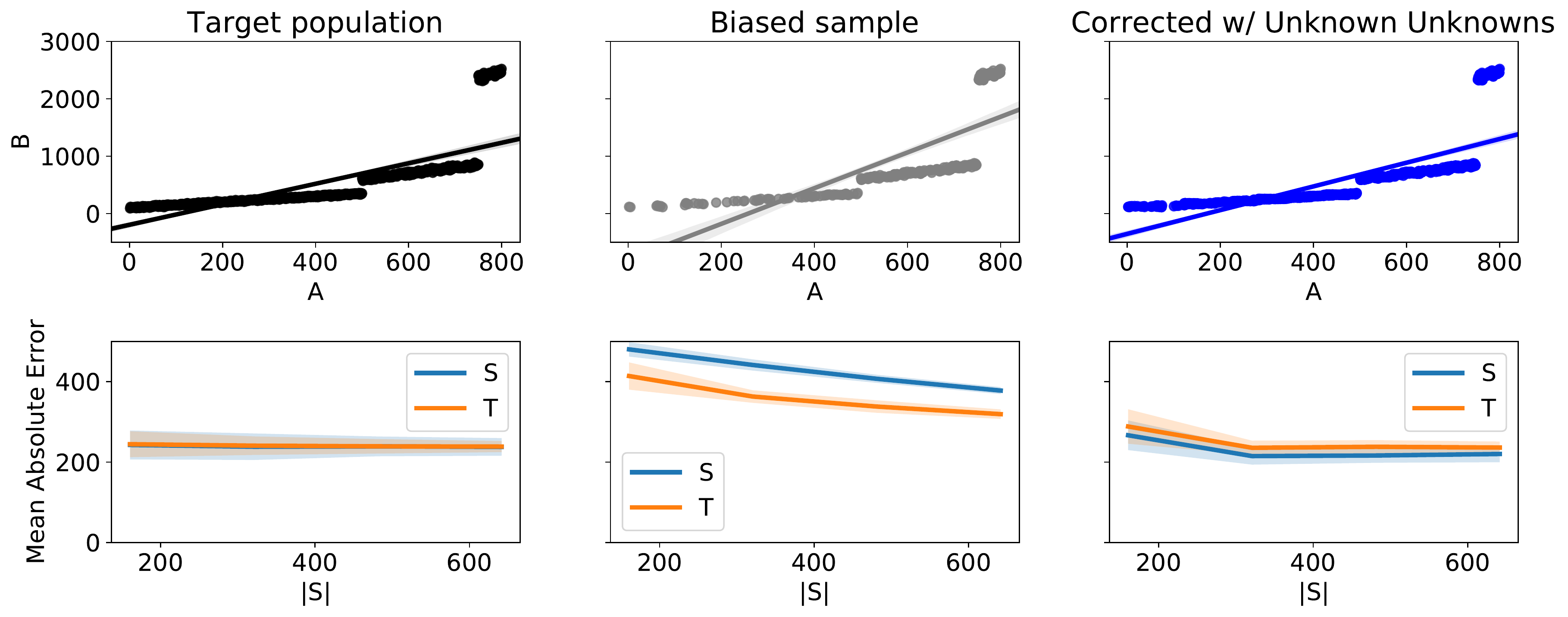}
  \caption{Ideally, we want the generalization gap between the training score ($S$) and the testing score ($T$) to be minimal (left), as in the ideal case--where we use a random sample from the underlying population. However, the model performs much worse if trained on a biased sample instead (middle) and fails to generalize to the actual testing data. Accounting for under-represented (or missing) data (a.k.a. \emph{unknown unknowns}) can improve the model generalizability (right)} 
  \label{fig:unknownml}
\end{figure}

Figure~\ref{fig:unknownml} illustrates the problem. In the toy example, the target population is hidden (only used for testing), but the training data, which is a biased sample from the population, is missing some of the examples with smaller $A$ values (e.g., smaller companies less likely to be sampled). The fitted regression model can still perform well on the training set, but will fail in testing. 


\subsection{Our goal and approach}

\begin{figure}[t]
   \centering
     \includegraphics[width=\columnwidth]{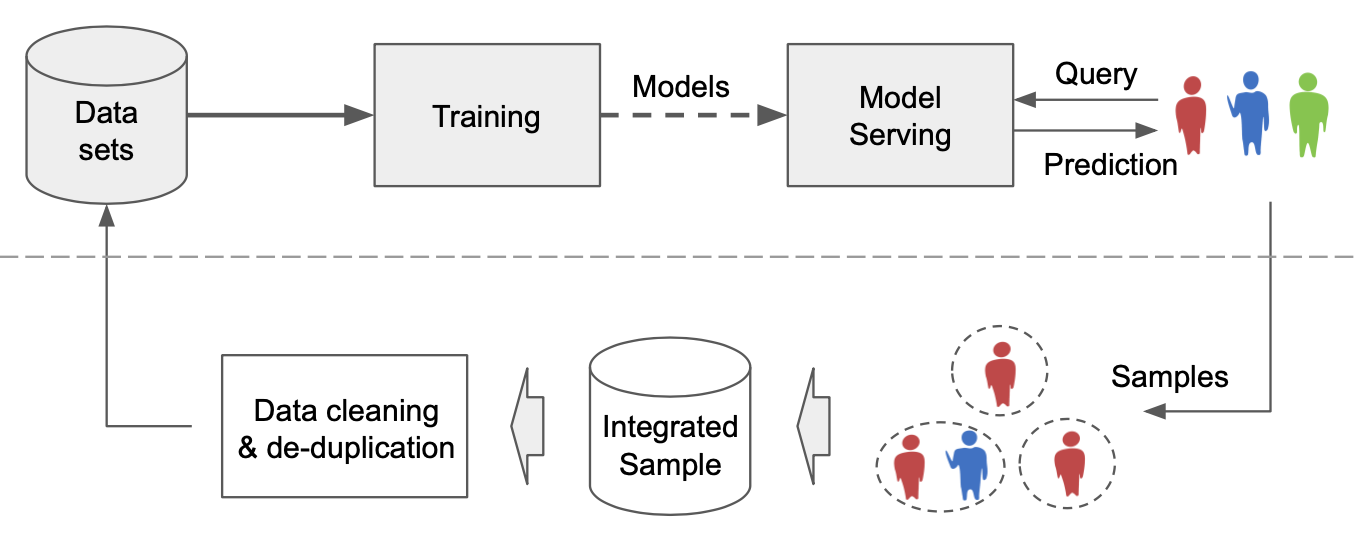}
     \caption{A typical ML pipeline (top) assumes a training dataset that is a representative random sample from the target population; however, the training data can be shifted or biased from the target population rather frequently due to a myriad of reasons. In this work, we leverage the fact that the training data is collected over multiple and redundant data sources, as in a typical data collection pipeline (bottom), to check and correct any covariate shfit between the training and the hidden testing data (target population).}
 \label{fig:ml_workflow}
\end{figure}

We aim to develop methods for mitigating unrepresentativeness in training data arising from sampling bias, or covariate shift more generally, thereby improving ML generalization performance. Our key idea is to exploit the fact that training data is typically created by integrating overlapping datasets, so that instances often appear multiple times in the combined data. Figure~\ref{fig:ml_workflow} depicts a sampling process where training data is collected over multiple and redundant data sources or samples, all sampling without replacement from the same population. Because a single data source is typically incomplete and has inconsistencies and data errors, such a sampling process---e.g., crowdsourcing \cite{ftlabel19}---is often employed in practice. We treat data cleaning as an orthogonal problem, as any proper data cleaning techniques can be applied without altering the problem context.

Given an integrated dataset that has not yet been de-duplicated, we first apply species-estimation techniques that use the multiplicity counts for the existing training instances to estimate the number of unknown instances. We then use this information to correct the sample by either weighting existing instances or generating synthetic instances. For the latter approach, we investigate both kernel density and interpolation techniques for generating feature values for the synthetic instances. As shown in our experiments over different types of ML models and datasets, correcting a training set by taking unknown instances into account can indeed improve model generalization.

\section{Related Work}

To the best of our knowledge, this is the first work to consider learning under covariate shift without an access to the (unlabeled) test data, but instead using species estimation techniques.

Learning under covariate shift or sample selection bias has been studied extensively \cite{bickel2007discriminative,sugiyama2013learning,zadrozny2004learning,liu2014robust,huang2007correcting,shimodaira2000improving}, because training and test distributions diverge quite often and for many reasons in practice.  
As mentioned in Section~\ref{sec:intro_covariate_shift}, most known learning techniques under covariate shift require unlabeled test data, which may not be available in many real applications.

Some recent studies~\cite{attenberg2011beat,lakkaraju2017identifying,bansal2018coverage} use the term "unknown unknowns" to refer to prediction errors having high reported confidence, which can arise due to any mismatch between the training data and the testing data~\cite{lakkaraju2017identifying}.
Instead, we define unknown unknowns, perhaps more appropriately, as unknown test examples that are unseen during the training (Definition~\ref{eqn:unknown_unknowns_ml}), which can arise due to covariate shift in real data collection scenarios (e.g., data integration over multiple data sources  \cite{chung2018estimating}).
More importantly, prior work on unknown unknowns \cite{attenberg2011beat,lakkaraju2017identifying,bansal2018coverage} assumes an oracle of true labels and tries to minimize the number of queries to it. In contrast, our problem assumes no such oracle or even any unlabeled testing data.

The situation where training and test data follow different distributions is also related to \emph{transfer learning}, \emph{domain adaptation} and \emph{dataset-shift adaptation} \cite{pan2010survey,daume2006domain,sugiyama2017dataset}. 
In transfer learning, a model built for one problem is applied to a similar problem. 
Here, the conditional distribution $p(y|x)$ is not constant and the learner is even asked to predict different labels. In that case, model is re-trained, at least partially, to adapt to a new problem.
Furthermore, transfer learning is often useful when the problem is complex (e.g., computer vision tasks) and requires a complex neural network model with lots of hidden layers. We mostly focus on simpler examples where it makes more sense to build the model from scratch.

There are also techniques to detect covariate shift. The most intuitive and direct approach would be take the two distributions, training and testing, and use the Kullback-Leibler divergence model~\cite{KullbackL51} or Wald-Wolfowitz test~\cite{wald1940test} to detect any significant data-shift.
Researchers also have looked at covariate shift detection where the distribution is non-stationary~\cite{raza2015ewma}. In our case, the testing distribution is hidden, so such comparisons are impossible.

Species estimation techniques have been studied in prior work for distinct count estimation, data quality estimation, and crowdsourced data enumeration \cite{DBLP:conf/vldb/HaasNSS95,Chung:2017:DQM:3115404.3115414,chung2018estimating,DBLP:journals/cacm/TrushkowskyKS16}. In this work, we use species estimation techniques to model unknown examples that are missing from the training data and not known to exist (Definition~\ref{eqn:unknown_unknowns_ml}). This allows us to correct biased training data without the test data.

\section{Unknown Examples \& ML Model Generalization}
\label{sec:unknownml:problem}

In this section, we define \emph{unknown unknowns} \cite{chung2018estimating} in the context of ML, and describe how common data collection procedures can produce a biased training data with unknown unknowns. Our goal is twofold. First, we want a model that performs well on unseen examples. Second, we want to minimize the gap between training and testing scores, so that the former will be truly predictive of the latter, and an ML model can be applied judiciously.

\subsection{Problem Setup}

A typical training data collection process involves sourcing and integrating multiple data sources, e.g., data crowdsourcing where each worker is an independent source \cite{hsueh2009data,lease2011quality}. In this work, we assume that data sources are independent but overlapping samples $S_j$, each obtained by sampling $n_j=|S_j|$ data items from the underlying distribution $p$; the sampling is without replacement, because a data source typically only mentions a data item once. $p$ is also our target distribution for learning, and each data item $x_i=<x_{i1}, x_{i2}, ..., x_{id}>$ has a sampling likelihood $p'(x_i)$ and consists of $d$ features/variables. The data sources are then integrated into a training data set $S$ of size $n_S = \sum_{j=1}^l n_j$. $S$ contains duplicates because every data source is sampling from the same underlying population. If we integrate a sufficiently large number of sources, then $S$ approximates~\cite{chung2018estimating} a sample with replacement from $p'$; we use this approximation throughout. The duplicate counts resulting from the overlap of the  $S_j$'s enables the use of species estimation techniques to estimate the number of the missing, unseen test instances $U$.

Ideally, we would like the integrated sample $S$ to follow the target distribution $p(x)$ for $x \in S$ (i.e., $S$ is a uniform random sample from $p$, like the hidden testing data $T$). However, we assume \emph{covariate shift} between $p(x)$ and the actual training-data distribution of $S$, denoted as $p'(x)$. That is, $p(x) \neq p'(x)$ but $p(y|x) = p'(y|x)$ for all $x$. As discussed previously, this situation may arise if (1) any of the integrated sources exhibits a strong sample selection bias or (2) the source is outdated, but the fundamental relationship between $x$ and $y$ is unchanged.

For training, we assume that, for $S=\{(x_i, y_i)\}$, each class label $y_i \in Y$ is perfectly curated. Neither testing data $T$---nor the distribution $p(x,y)=p(y|x)p(x)$ that generates $T$---is available during training. We assume that the hidden test data comprises an i.i.d.\ sample from $p(x,y)$, and well represents this distribution.

\subsection{The Unknown Examples}

We focus on the missing training examples that actually exist in the test set, and now formally define such missing examples as \emph{unknown unknowns}.
We write $V\sim p$ to indicate that $V$ is obtained by repeated sampling with replacement according to probability distribution $p$, followed by removal of duplicates. We assume that the probability of sampling a given item depends solely on the $(x,y)$ attribute values of the item, so that $p$ can be viewed as a distribution over the attribute space of the population. 

\vskip 0.5\baselineskip
\begin{definition}[Unknown Unknowns]
Let $S \sim p'(x,y)$ be an integrated sample for training and $T\sim p(x,y)$ be a (hidden) sample for testing, both sampled from the same population (i.e., sample space). The set of \emph{unknown unknowns} $U$ is then defined as $U=T-S$.
\label{eqn:unknown_unknowns_ml}
\end{definition}
\vskip 0.5\baselineskip

Note that, by definition, we will have unknown unknowns even when there is no  covariate shift ($p=p'$). Moreover, in general, approximating unknown unknowns can both improve or harm model quality \cite{Ha:1997:OHN:252862.252903}. Our experiments indicate, however, that if a user applies our techniques (not knowing whether there is covariate shift or not), degradation of model generalization performance tends to be small at worst, and the improvement in quality when $p\neq p'$ is typically significant.


The terminology ``unknown unknowns" stems from the fact that both the cardinality of $U$ and the feature values of $x\in U$ are unknown. The existence of unknown unknowns critically impacts a model's generalization ability. 

\subsection{Problem Statement}
\label{sec:unknownml:problem:statement}
 We quantify a model's generalization ability via \emph{generalization error}, the difference between the error (expected loss) with respect to the underlying joint probability distribution and the error (average loss) on the finite training data. That is,
 \[
G  = \int L(f(x), y)p(x, y)dxdy - \frac{1}{n_S} \sum_{i\in S} L(f(x_i), y_i),
\]
which we approximate by the empirical generalization error
\[
G_e=\frac{1}{n_T} \sum_{i\in T} L(f(x_i), y_i) - \frac{1}{n_S} \sum_{i\in S} L(f(x_i), y_i).
\]
Here $n_A=|A|$ and $L$ is a loss function such as $L(x,y)=(x-y)^2$.

Our goal is to minimize generalization error
in order to maximize the predictive ability on new data. 
However, $\sum_{i\in T} L(f(x_i), y_i)$ is not available during training, and so ML training algorithms aim to minimize the empirical risk $(1/n)\sum_{i\in S} L(f(x_i), y_i)$, where $(x_i, y_i)\sim p'(x,y)$.
Thus, any significant discrepancy between $p(x)$ and $p'(x)$ will be reflected in $G$ through the average loss over the unknown examples. Training on $S$ when $|U| >> 0$ can result in poor model performance on the actual testing data.

We now define the impact of unknown unknowns in the context of the generalization error.
\vskip 0.5\baselineskip
\begin{definition}[The Impact of Unknown Unknowns]
\label{prob:unknown_impact}
Given an integrated data set $S\sim p'(x,y)$ for training and a (hidden) testing data set $T\sim p(x,y)$, the impact of unknown unknowns is defined as 
$\Delta = (1/n_U) \sum_{i\in U} L(f(x_i), y_i)$.
\end{definition}
\vskip 0.5\baselineskip
Some straightforward algebra shows that
\[
G_e = \frac{n_U}{n_T}\Delta+\big(\frac{1}{n_T} - \frac{1}{n_S}\big)\sum_{i\in S} L(f(x_i), y_i).
\]
For given sample sizes, $n_T$ and $n_S$, and thus $n_U$, $G_e$ is an increasing affine function of $\Delta$, so that decreasing $\Delta$ will decrease $G_e$. In Section~\ref{sec:unknownml:learning}, we explain how adding $U$ to $S$ can correct the training distribution $p'(x)$ to resemble $p(x)$, which in turn will decrease $\Delta$.


Our definition of unknown unknowns (Definition~\ref{eqn:unknown_unknowns_ml}) more naturally fits the selection-bias model~\cite{zadrozny2004learning} where $S$ is sampled from $T$ (so that $S\subseteq T$), and the definition only concerns missing testing instances from $T$, not additional training instances in $S-T$.
In the most general setting, where training and testing data can differ arbitrarily, the additional training instances in $S-T$ can potentially help or hinder model accuracy.
We leave analysis of the general setting as future work.


The goal is to estimate the distributional difference between $p(x)$ and $p'(x)$ via unknown unknowns, and make the unknown examples part of model training.
The challenge arises because neither $T$ nor $p(x)$ is available at training time; we cannot directly compute $U$.  

\section{Learning The Unknown}
\label{sec:unknownml:learning}
In this section, we focus on a simple regression problem to illustrate our techniques for learning the unknown examples. 
The proposed techniques can easily be extended to other problem types, such as classification (apply the same technique for each class label). In Section~\ref{sec:experiment}, we present the experimental results for both regression and classification problems.

\begin{figure}[t]
    \centering
        \includegraphics[width=\columnwidth]{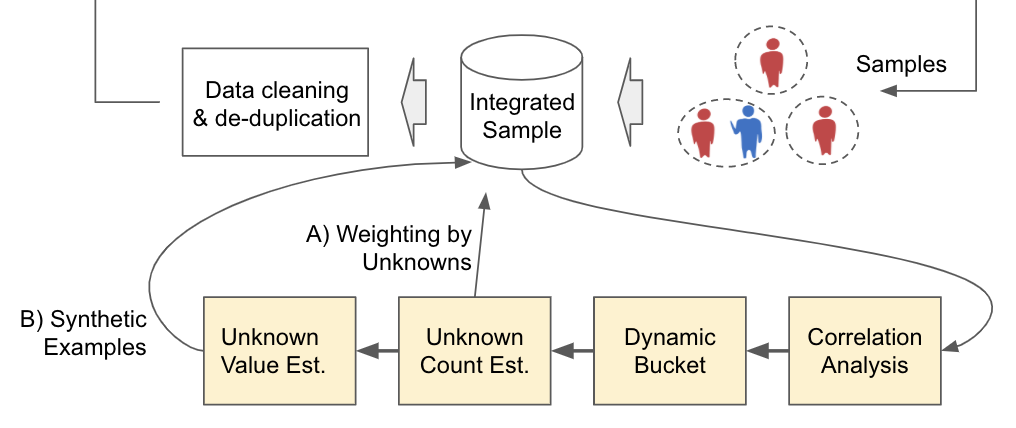}
\caption{We propose two approaches for learning the unknown: A) weighting by unknown example count; B) generating synthetic unknown examples.}
\label{fig:unknown_learning_overview}
\end{figure}

We propose two approaches to model the unknown examples using the integrated sample with duplicates. As mentioned previously, the key idea is to use information about duplicate values in the integrated training data to estimate the missing unknowns. Figure~\ref{fig:unknown_learning_overview} outlines the following steps to model the unknown examples.

\subsubsection{Correlation Analysis} We first select the feature most correlated to the dependent variable $y$, and use it as the basis for the unknown count estimation. This is important because the species estimation technique we use for unknown count estimation is a univariate estimator, and  unknown unknowns count estimates can vary, depending on the feature we choose; we want to use the most relevant feature for the final ML task.
Alternatively, we can choose the feature with the highest entropy or variance.

\subsubsection{Dynamic Bucketization} We partition the data based on the values of the selected feature and perform the subsequent unknown count and value estimations in each partition. 
The more fine-grained the partition, the more accurately we can estimate the unknown example values, but the less accurately we can estimate the number of unknowns, since each partition contains fewer examples for count estimation. We use the bucketization algorithm described in Section~\ref{sec:dynamic_bucket}.

\subsubsection{Unknown Count Estimation} For each data partition, we estimate the number of unknown examples using \emph{Chao92} species estimation technique as outlined in Section~\ref{sec:species_estimator}. We can use the count estimates to re-scale existing training examples; all examples within a partition get the same weight, namely, the number of unknown examples in that partition. The assumption is that the higher the unknown example count in a partition, the more likely that the corresponding region from the population is under-represented in the training sample.

\subsubsection{Unknown Value Estimation.} As an alternative to the weighting scheme, we can model the unknown unknowns more explicitly by estimating the unknown examples' (feature) values. Generating meaningful examples is not straightforward and we use techniques described in Sections~\ref{sec:kde_value_estimator} and \ref{sec:smote_value_estimator}.

\subsubsection{Learning.} The modified dataset is provided to the learner after the usual data processing steps, e.g., transformation, feature engineering, data cleaning, de-duplication, and so on.

We now discuss each of the proposed approaches in more detail.

\subsection{Weighting by Unknown Example Count}
\label{sec:unknown_weighting}
The first approach simply weights an existing training example by the estimated number of unknown unknowns that surround it. This is similar to the importance-sampling-based techniques studied previously, where the instance-specific weights approximate $p(x)/p'(x)$; see \cite{shimodaira2000improving,bickel2007discriminative,zadrozny2004learning}. Other techniques exist for learning the weights, directly or indirectly, from the training and testing distributions \cite{huang2007correcting,liu2014robust}, but they are not applicable if the test data $T$ or the ground truth distribution $p(x)$ is not available. 

To this end, We use a sample-coverage-based species estimation technique to estimate the size of the missing mass. The underlying assumption of the estimator is that the rare species in a sample with replacement are the best indicators of the missing unknown species in the target distribution.
In our case, we assume that the rare examples in our collected $S$ are the best indicators of the unknown unknowns. If we estimate a large number of unknown unknowns near a given training example, then we expect a large number of similar instances to be present in the actual test data or target population. Thus, we attach more importance to such a training example during the training phase. 

\subsubsection{Chao92 Species Estimator}
\label{sec:species_estimator}
Because neither the test data $T$ nor the ground truth distribution $p(x)$ is available, we use a sample-coverage-based species estimation technique to estimate the number of missing unknown examples. A sample-coverage-based estimation scheme looks at the size of the overlaps in the collected sample to reason about the existence of the unknown missing mass from the population. The intuition is that we are less likely to uncover new species if we keep sampling known species over and over again; conversely, if many species appear in the sample only once, then there are likely more species that we have not yet seen.

There are several species estimation techniques to estimate the unknown (distribution) mass, and no single estimator performs well in all settings~\cite{DBLP:conf/vldb/HaasNSS95}. We estimate the true number of values $|T|$ using the popular \emph{Chao92} estimator, which is defined as
\[
\hat{D}_{Chao92} =(c/\hat{C}) + (f_1\cdot\hat{\gamma}^2/\hat{C}),
\]
where $\gamma$ is \emph{coefficient of variation} and can be estimated as:
\[
\hat{\gamma}^2 = \max \left\{\frac{\frac{c}{\hat{C}} \sum_i{ i(i-1)f_i}}{n(n-1)} - 1 \, , \,0\right\}
\]
Here $c$ is the number of unique examples in the training data $S$, $\hat{C}$ estimates the sample coverage $C$---i.e., the percentage of $T$ covered by $S$---and $D_{Chao92}$ our estimate of the total number of unique examples in $T$. For $i\ge 1$, the quantity $f_i$ denotes the number of examples that occur exactly~$i$ times in the integrated training data $S$ (before duplicate removal).
The sample coverage is estimated using the Good-Turing estimator~\cite{turing53}:
$\hat{C} = 1 - f_1/n$;
here $f_1/n$ estimates the missing distribution mass of the unknown unknowns.

\subsubsection{Dynamic Bucketization}
\label{sec:dynamic_bucket}
To estimate the number of unknown unknowns near each training example, we partition the data into buckets based on the values of a selected feature and then perform the estimation for each partition. Instead of partitioning the feature space statically, with fixed boundaries and sizes, we define the buckets dynamically, making sure that each partition contains enough examples and duplicates to permit high-quality estimation.

\begin{algorithm}[ht]
\small
\SetAlgoLined
\SetKwInOut{Input}{Input}\SetKwInOut{Output}{Output}
\Input{Integrated training data $S$, feature index $v$, min sample coverage threshold $\theta$}
\Output{data partitions (buckets) $B$}
\BlankLine
    $B=[]$\tcc*[r]{buckets}
    $Q=PriorityQueue(S)$\tcc*[r]{priority queue sorted by ascending feature value $x_{iv}$ of $x_i\in S$}
    $b=[]$\tcc*[r]{current bucket to fill}
    \BlankLine
    \While{$Q$ not empty}{
      $x = Q.pop()$\tcc*[r]{the next $x_i$ by ascending $x_{iv}$}
      \If{sample\_coverage($b$) $\geq \theta$}{
        $B.append(b)$\tcc*[r]{if $\hat{C} \geq \theta$, $b$ has enough}
        $b=[x]$\tcc*[r]{new bucket to fill}
      }
      \Else{
        $b.append(x)$\;
      }
    }
    \BlankLine
    return $B$\;
    \caption{Dynamic Bucketization}
    \label{alg:dynamic_buckets}
\end{algorithm}

Algorithm~\ref{alg:dynamic_buckets} illustrates the mechanism. First, we push the training data $S$ onto a priority queue sorted by ascending feature value $x_{iv}$ (line 2). The feature index $v$ is selected based on the feature correlation to the class label $y$, in order to estimate the counts with respect to the most informative feature. 
Our strategy ensures that we do not scale examples by missing values in a less relevant feature dimension for the final ML task.
We plan to explore other metrics and algorithms for the feature selection in the future.
Afterwards, we group nearby examples $x_i\in S$ in a way that each bucket has enough examples and duplicates, according to the sample coverage estimate, for the quality unknown unknowns species estimation (lines 4-13).

Notice that Algorithm~\ref{alg:dynamic_buckets}  greedily defines buckets simply to ensure every bucket has enough data for unknown unknowns species estimation.
Optimizing for a global objective (e.g., the best model performance on a training dataset) is meaningless unless we assume there is no sampling bias or covariate shift. The challenge is that the testing data (and the underlying target distribution) is unknown and not available in any form (e.g., labeled/unlabeled testing data or a validation dataset as a representative sample of the actual testing data).
We found that the algorithm and the equal-sized bucket with minimum sample-coverage threshold works well, but it would also be interesting to explore other hierarchical clustering algorithms \cite{steinbach2000comparison} to exploit data structure in the feature space.

\begin{example} [Weighting by Unknown Example Count]
Consider a training set $S=\{(x_1,y_1), (x_2, y_2), ... , (x_{1000},y_{1000})\}$ of size 1000 with 2 features, $g_1$ and $g_2$. Supposing that, over $S$, feature $g_2$ is more highly correlated with the class label than $g_1$, we use $g_2$ to  partition the training data into buckets $\{b_1, b_2, b_3\}$ using the dynamic bucketization algorithm. Each bucket $b_i$ has a sample coverage greater than $\theta=0.5$, which enables accurate unknown unknowns estimation per bucket. If, e.g., there are more unknown unknowns in the feature-space region covered by $b_2$ than those of $b_1$ and $b_3$, then we place more importance on---i.e., give more weight to---the examples in $b_2$. Given the unknown unknowns count estimates of 150 for $b_1$, 400 for $b_2$ and 50 for $b_3$, our unknowns learner assigns a weight of 150/600 to all the examples in $b_1$, 400/600 to all the examples in $b_2$ and 50/600 for $b_3$. It then re-trains the ML model (e.g., logistic regression classifier) using the weighted training instances.
\end{example}

\subsection{Synthetic Unknown Examples}
\label{sec:unknown_injecting}

The second approach tries to model the unknown unknowns more explicitly and generate synthetic training examples. That is, for each bucket $b$ obtained using dynamic bucketization (Algorithm~\ref{alg:dynamic_buckets}), we use the species estimation technique to estimate a number $n_b$ of unknown unknowns and then generate $n_b$ synthetic unknown unknowns.
This approach might appear more risky than the weighting approach in Section~\ref{sec:unknown_weighting}, because it requires estimation of feature values, a potential source of additional uncertainty and error. Our experiments show, however, that this approach can be very effective, if done carefully.

A na\"ive approach would use \emph{mean substitution} \cite{schafer2002missing}, where the average observed feature values are used for any unknown unknowns. We could also try doing this at the bucket-level \cite{chung2018estimating}, but all in all, we have found that this does not add much value to the learner. Furthermore, the weighting approach does not work for some ML algorithms that cannot train on weighted training data, e.g., nearest neighbor classification or neural networks.
Instead, we use a couple of data-driven oversampling techniques for unknown unknowns value estimation that are more aggressive than the weighting approach, but also do not generate values that depart arbitrarily far from the observed data distribution $p'(x)$ (e.g., long tail distribution).
It is important that we use these conservative data-driven approaches to avoid adding bad examples and outliers that can harm the model's generalization performance.
In the following, we describe two estimators for unknown feature values.

\subsubsection{KDE-Based Value Estimator.}
\label{sec:kde_value_estimator}
We use a Kernel Density Estimation (KDE) approach~\cite{silverman1986density} to estimate the probability density of each bucket and sample the missing unknown examples from it. This is effective, especially when covariate shift is mainly due to sample selection bias, i.e., $S\subset T$, and unknown unknowns are similar to the observed training examples.
On the other hand, the value estimator can actually mislead the training if $p'(x,y)$ is very far apart from $p(x,y)$.
We used a Gaussian kernel and a ``normal reference'' rule of thumb \cite{henderson2012normal} to determine the smoothing bandwidth, but other kernels are possible, and a subject for future study.

\subsubsection{SMOTE-Based Value Estimator}
\label{sec:smote_value_estimator}
Synthetic Minority Oversampling Technique (SMOTE) is widely-accepted technique to balance a dataset \cite{chawla2002smote}. A dataset is said to be imbalanced if different class examples are not equally represented. SMOTE generates extra training examples for the minority class in a very conservative way; the algorithm randomly generates synthetic examples in between minority class examples and their closest neighbors. Motivated by this class-label-balancing algorithm, we generate synthetic unknown examples in a similar fashion.

\begin{algorithm}[h]
\small
\SetAlgoLined
\SetKwInOut{Input}{Input}\SetKwInOut{Output}{Output}
\Input{Integrated training data $S$, expected number of unknown examples $l$, number of nearest neighbors $k$}
\Output{Synthetic unknown examples $U$}
\BlankLine
    $U=[]$\tcc*[r]{synthetic unknown examples}
    
    \BlankLine
    \For{$i=1,2,\ldots,l$}{
         $x_i=random(S)$\tcc*[r]{randomly pick $x_i \in S$}
         $N = kNN(x_i, S, k)$\tcc*[r]{$k$ nearest neighbors of $x_i$}
         $x_j=random(N)$\tcc*[r]{randomly pick $x_j \in N$}
         $u=init(d)$\tcc*[r]{synthetic example with $d$ features}
         \BlankLine
         \For{$f=1,2,\ldots,d$}{
            $\delta = x_{jf} - x_{if}$\;
            $g = random(0, 1)$\tcc*[r]{random number b/w 0 and 1}
            $u[f] = x_{if} + g*\delta$\tcc*[r]{generate new feature value}
         }
         \BlankLine
         $U.append(u)$\;
     }
    \BlankLine
    return $U$\;
    \caption{SMOTE-Based Value Estimator}
    \label{alg:smote}
\end{algorithm}
 
Algorithm~\ref{alg:smote} illustrates how the synthetic unknown examples are generated.
We generate exactly $l$ unknown examples, where $l$ is computed using the species estimation technique (line 2). We first initialize a dummy synthetic example with $d$ features (an arbitrary example from the same feature space (line 6). Next, we take $k$ nearest neighbors $N$ of a randomly picked example $x_i\in S$, and also pick a neighbor $x_j\in N$. Setting $k$ high results in more aggressive data generation, since $\delta$ (line 8) can be larger. Finally, we randomly generate $d$ features values by randomly interpolating between $x_i$ and $x_j$ (lines 7-11).

\section{Experiments}
\label{sec:experiment}

We designed our experiments to explore (i) how the proposed techniques compare to each other, (ii) how the techniques compare to the prior work that requires unlabeled testing data, and (iii) whether learning-the-unknowns techniques improve model generalization on real crowdsourced datasets. 

\subsection{Experimental Setup}

\begin{figure*}[h]
  \centering
     \includegraphics[width=\textwidth]{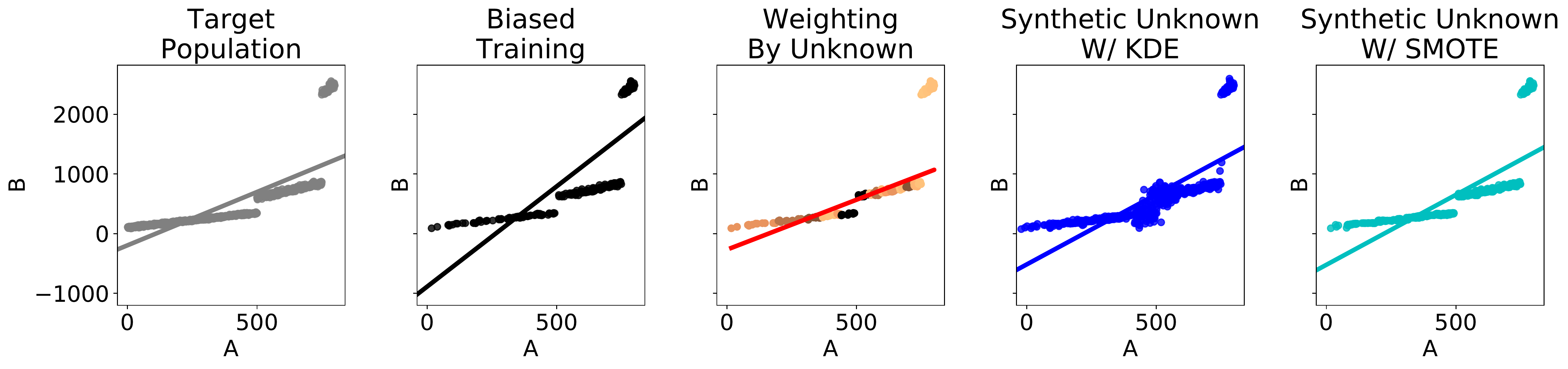}
     \caption{From left to right, we have a target population $T$ and the optimal linear regression model (gray line), a biased training data $S$ (in this case, examples with smaller values for the dependent variable $A$ are less likely to be sampled) and the fitted model (black line), weighted training examples by the unknown count estimates (darker colors correspond to higher weights) and the model (red line), and enriched training data sets with two different kinds of synthetic unknown examples and their fitted models (blue and cyan lines). }
 \label{fig:toy_regression_overview}
\end{figure*}

We evaluated our unknown-example learning techniques using both simulated and real-world crowdsourced datasets. 

For the simulated-data experiments, we used datasets from UCI Machine Learning repository\footnote{https://archive.ics.uci.edu/ml/datasets.html} as base population datasets; we re-sampled multiple times from each base dataset with sample selection bias, and then combined the samples, yielding a biased training dataset (with duplicates). A random uniform sample from the base dataset was taken and hidden for use as a testing dataset.

For the real-world datasets, we used Amazon Mechanical Turk (AMT) for crowdsourcing. We paid $\$0.03$ for each HIT, i.e., for each example or data item. We collected responses over multiple workers,  treating each worker as an independent data source. For each question posed---e.g., ``What are the heights and weights of active NBA players from 2015 to 2018?''---we integrated worker responses to get a training dataset. We expected to see some inherent sampling bias in workers' data collection processes, and thus, in the final training data. For each crowdsourced question, we have available the ground truth dataset, obtained from other sources. 

For each experiment, we compared the test accuracies of a specified classification or regression ML model that was learned over several different training sets: the original biased training data [\emph{\textit{Original}}], training data weighted as in  Section~\ref{sec:unknown_weighting} [\emph{\textit{WeightByUnk}}], and training data enriched with unknown unknowns using the two different value estimators from Section~\ref{sec:unknown_injecting} [\emph{\textit{SynUnk(KDE)}} and \emph{\textit{SynUnk(SMOTE)}}]. We also present an ideal case where the model is trained on the testing data [\emph{\textit{Ideal}}]; this is similar to using techniques that query an oracle~\cite{lakkaraju2017identifying}, but with an infinite amount of resources.
We use \emph{mean absolute error (MAE)} as our accuracy metric for regression problems:
\[
MAE = \frac{1}{m} \sum_{i\in T} |y_i - f(x_i)|
\]
and \emph{accuracy} for classification problems:
\[
ACC = \frac{TP+TN}{TP+FP+TN+FN}
\]
In binary classification with positive (true positive, TP, or false positive, FP) and negative (true negative, TN, or false negative, FN) class labels, accuracy measures the fraction of correctly predicted instances.
The scales on the y-axes are not normalized and dependent on the actual target variables for the problems.

Figure~\ref{fig:toy_regression_overview} provides an overview of different techniques and their results using the example from Figure~\ref{fig:unknownml}. For this toy example, it can be seen that estimating unknown unknowns results in regression lines that resemble the true regression line (leftmost plot) much more closely than does the line based on the original biased sample.

\subsection{Simulation Study}
We use the following public ML datasets for the simulation study:

\textbf{Auto MPG Dataset \cite{quinlan1993combining}.} The task is to predict city-cycle fuel consumption in miles per gallon. There are 8 features, both numerical and categorical. The original dataset is collected over three different cities; to simulate sample selection bias, we sampled examples mostly from city 1 for training and use the examples from all the cities (1, 2, and 3) for testing. Training data was sampled with replacement to simulate a data collection process that combines multiple data sources.

\textbf{Image Dataset \cite{kaggle_dog_cat}.} An image classification problem using a set of 25K cat and dog images, re-sampled and split into biased training data and unbiased testing data. We inject a sampling bias so that sematically meaningful sub-groups are missing from the dataset \cite{lakkaraju2017identifying}; the training dataset comprises of black dog and non-black cat images, whereas the testing contains all colors of dog/cat at random. We use this dataset to compare the proposed techniques with existing techniques. We also set the cat class label to be the critical class in the experiment.

For all simulations, we permuted/re-sampled the training dataset to repeat the experiments $r=20$ times. 

\begin{figure*}[t]
   \centering
     \includegraphics[width=\textwidth]{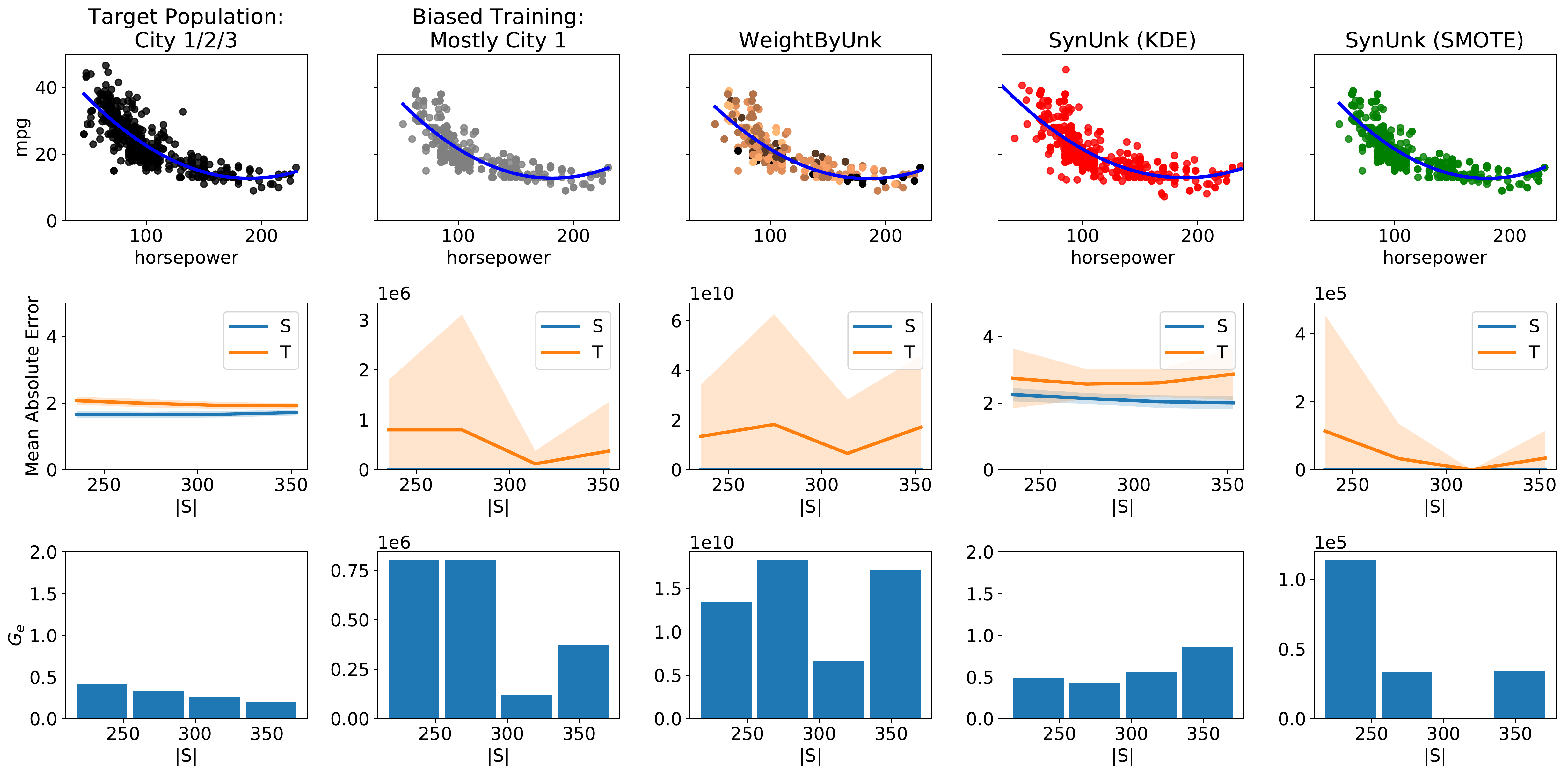}
     \caption{UCI Auto MPG Dataset model evaluation error on training $S$ and testing $T$ and the impact of unknown unknowns (generalization error)--please notice the scientific notation on some of the y-axes.The leftmost column (Target: City 1/2/3) is an ideal case where we train on $T$ ($S=T$); and we have a biased training case where the data mostly consists of records from City 1 in the middle column. We see that SynUnk(KDE) improves the model generalization both in terms of testing performance (mean absolute error, the lower the better) and generalization error, $G_e$, in the rightmost column.}
 \label{fig:auto_mpg}
\end{figure*}

Figure~\ref{fig:auto_mpg} illustrates how our techniques can reduce the generalization error and the generalization error gap between the training and the testing scores
 We see that SynUnk(KDE) dramatically improves the
  regression model generalization both in terms of testing performance (lower mean absolute error) and generalization error, $G_e$, compared to the biased training case.
This means that the training error better represents what is to be expected on the actual testing data. This is desirable in practice, since ML algorithms assume that the testing  distribution will follow the training and therefore optimize for the best training score/error. Under covariate shift this assumption can be problematic.


For this experiment, WeightByUnk has the worst accuracy and largest generalization error. This is somewhat unexpected, since WeightByUnk does not require estimating the unknown features. SynUnk(SMOTE), which estimates unknown values but in a very conservative manner, also did not do very well. It appears that, for this example, the ability to extrapolate beyond observed values greatly helps the model's performance.

 
Next, we consider ML algorithms for classification.
The Adult dataset was divided into disjoint training and testing sets. To simulate a data collection pipeline with multiple sources and overlapping samples, we re-sampled the training set multiple times and combined the samples. The re-sampling mechanism was slightly biased to favor people with higher education backgrounds.
We also considered two existing techniques based on approximations to importance-sampling weighting~\cite{shimodaira2000improving}. \emph{2-Stage LR} first learns a logistic regression model $\hat{f}$ to classify if an example belongs to $S$ and $T$---this requires an access to $T$---and the scale-factor is then approximated as follows~\cite{bickel2007discriminative}:
\[
\frac{p(x)}{p'(x)} \approx \frac{n_S}{n_T} \cdot \left( \frac{1}{\hat{f}(x) }- 1 \right)
\]
$\hat{f}$ outputs the likelihood that $x$ belongs to $S$.
\emph{2-Stage LR (SSB)} is simpler in that it uses (normalized) $1/\hat{f}(x)$ to scale $x$. 
Both techniques look at $p(x)$ and approximate $p(x)/p'(x)$ to re-scale the original training data.
It is important to note that we make use of the testing data to run 2-Stage LR and 2-Stage LR (SSB), but not for our proposed techniques.

\begin{figure}[t]
  \centering
     \includegraphics[width=\columnwidth]{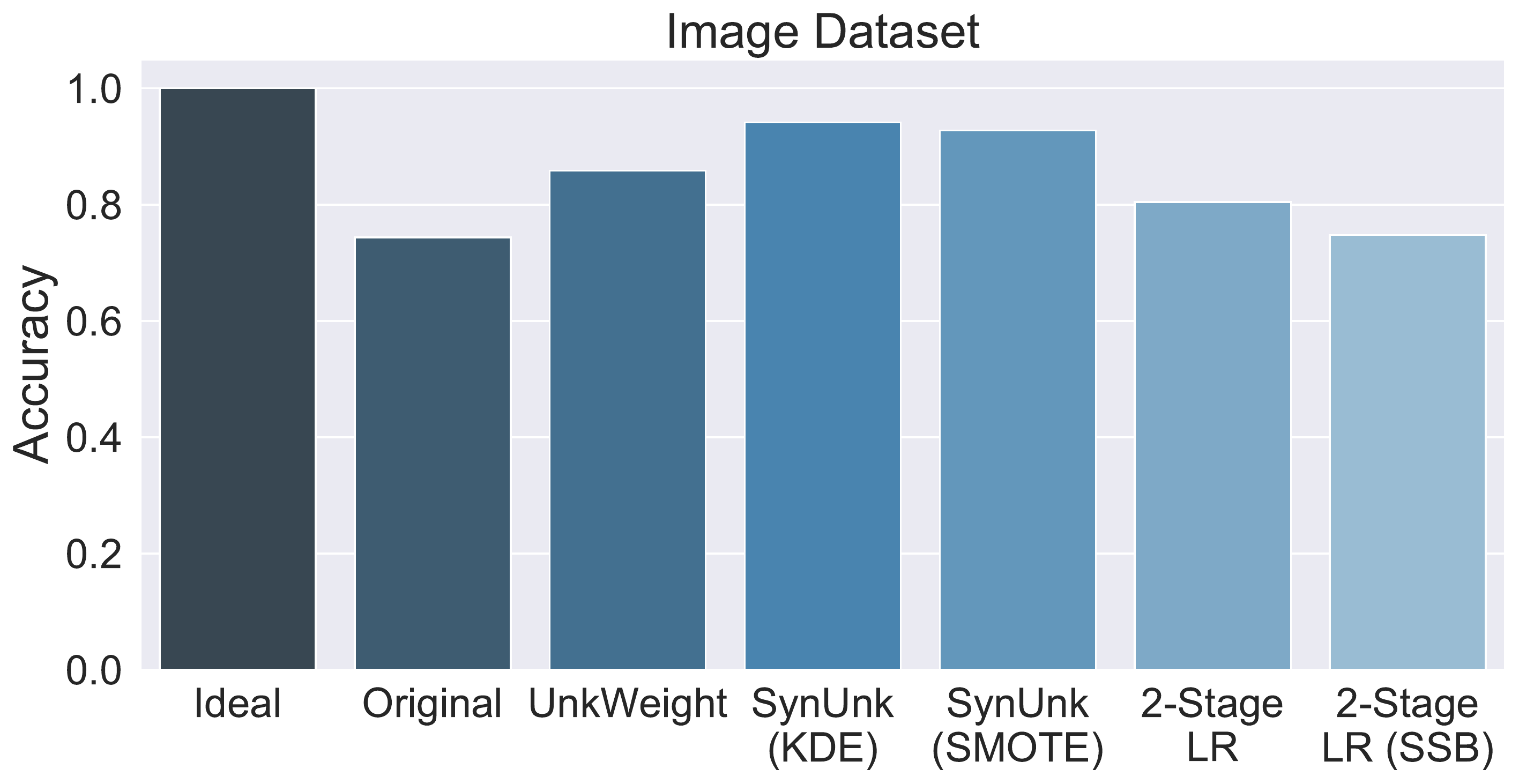}
     \caption{Image Dataset classification accuracy. We compare our unknown learning techniques to existing techniques: Ideal (\cite{lakkaraju2017identifying} with full access to the oracle), 2-Stage LR and 2-Stage LR (SSB). (\cite{bickel2007discriminative}).}
 \label{fig:image}
\end{figure}

\begin{figure}[t]
  \centering
     \includegraphics[width=\columnwidth]{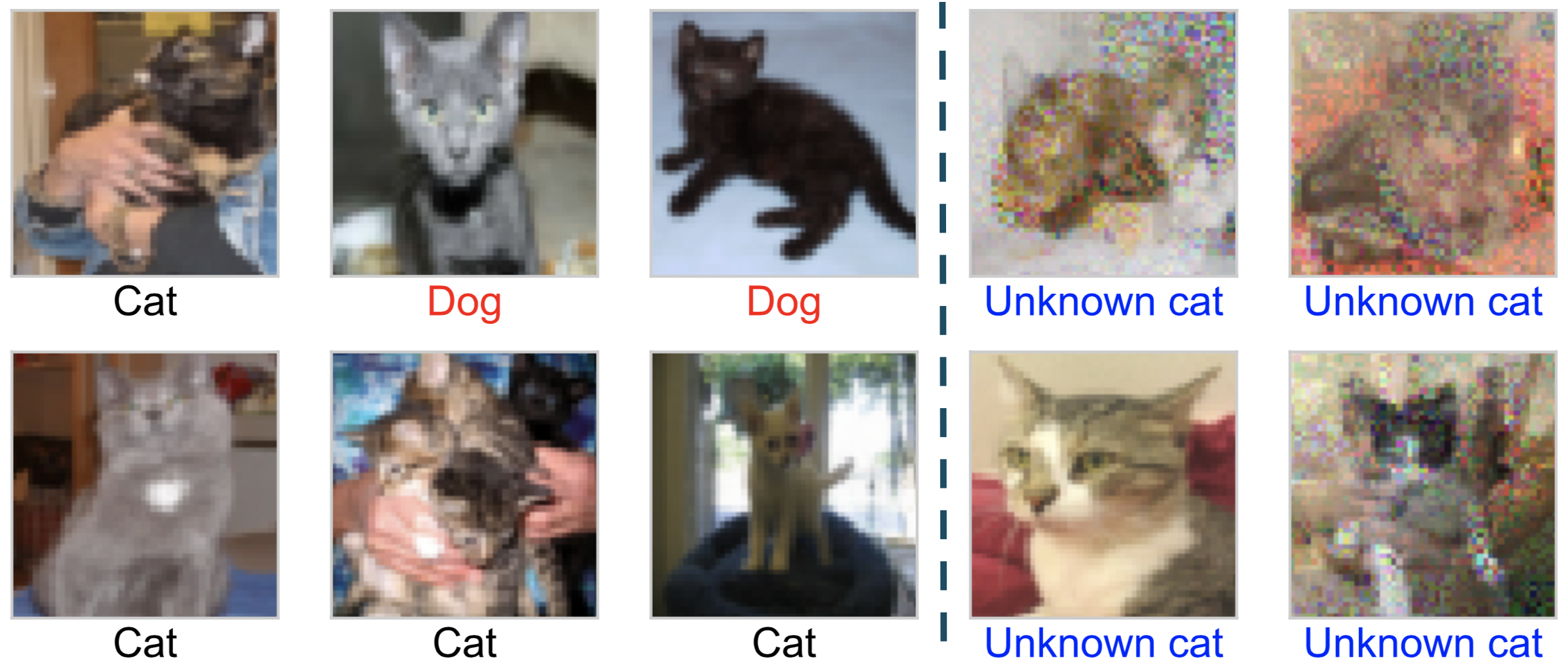}
     \caption{Examples where the original biased model mis-classifies some black cats as dog (left); Examples of synthetic unknown (cat) examples. The blur is expected as we generate unknown examples by smoothing (i.e., blurring) existing data points.}
 \label{fig:image_sample}
\end{figure}

Figure~\ref{fig:image} shows the image classification accuracy on the testing data using the proposed unknown unknowns techniques with a deep learning model comprising a two-layer dense network with ReLU and Softmax activation functions, trained with the Adam optimization algorithm~\cite{Kingma2014AdamAM}. Unless we have a full access to an oracle or the testing data for training (Ideal), the systematic bias in the training data can degrade the model generalization and performance (Original). Existing techniques, such as 2-Stage LR, leverage unlabeled testing data for importance-sampling weighting of the training data; we see only a slight improvement over the original model because assigning more importance to rare examples does not save the model from this extreme systematic bias, where there are almost no black cats to claim the importance. Similarly, the improvement we see in UnkWeight is also limited. On the contrary, we see that injecting unknown unknowns works well (SynUnk), unless it is random (Random).

In general, adding synthetic examples for image classification can provide a greater benefit if plausible data transforms is known (e.g., rotating/shifting images) \cite{wong2016understanding}, improving performance and reducing overfitting (better generalization). 
Species estimation with dynamic bucketization (SynUnk) works to identify the correct regions in the feature space with high/low unknown unknowns concentration. And by generating synthetic examples around regions with high unknown unknowns concentration using a conservative data-driven feature value estimation (e.g., SMOTE), the proposed unknown learning techniques try to improve the model generalization in case there is a systematic bias in the training data. Figure~\ref{fig:image_sample} shows the re-constructed images of the synthetic unknown unknowns examples.

It is also interesting to note that we picked the most correlated feature to the class labels to perform dynamic bucketization and unknown count estimation. In image classification, it makes sense to leverage the locality of the features (e.g., a patch of pixels) by using a convolutional neural network or super-pixels \cite{ribeiro2016should, lakkaraju2017identifying}. This requires adapting our techniques to multi-dimensional feature space directly, rather than picking a feature dimension to work with, and we plan to explore this for the future.

\subsection{Real-World Crowdsourced Examples}

We used AMT to collect two real-world datasets. Because we combined data from multiple crowd workers, our final training dataset contains redundant samples.
Workers sample from the same real-world universe for each question, and they are assumed to act independently; each worker provides an independent sample without replacement, and the combined sample contains duplicates to enable species estimation~\cite{chung2018estimating, DBLP:journals/cacm/TrushkowskyKS16}. Yet the combined dataset might still be biased---i.e., not a uniform random sample from the population---due to inherent selection bias of each worker; e.g., a worker might identify popular items while missing unpopular ones. 
The datasets are as follows.

\textbf{NBA Player Body Measurement.} To determine the relationship between height and weight, we crowdsourced body measurements of active NBA players. The final dataset $S$ contains 471 records, many of them redundant (more measurements collected for popular players). We have ground truth data for all 439 active NBA players, which we hide during training.

\textbf{Hollywood Movie Budget \& Revenue.}
The goal is to predict the gross revenue of a movie based on its production budget.
We used the crowd to collect production budget and world-wide gross revenue information (300 records with duplicates) for movies released from 1995 to 2018. The scale of movie production and gross revenue have changes a lot over the years; we used Hollywood movies released between 1995 and 2015 as test data, which should result in covariate shift between the training and testing data. As with the NBA example, we have ground truth data.


\begin{figure}[t]
\centering
\includegraphics[width=0.24\textwidth]{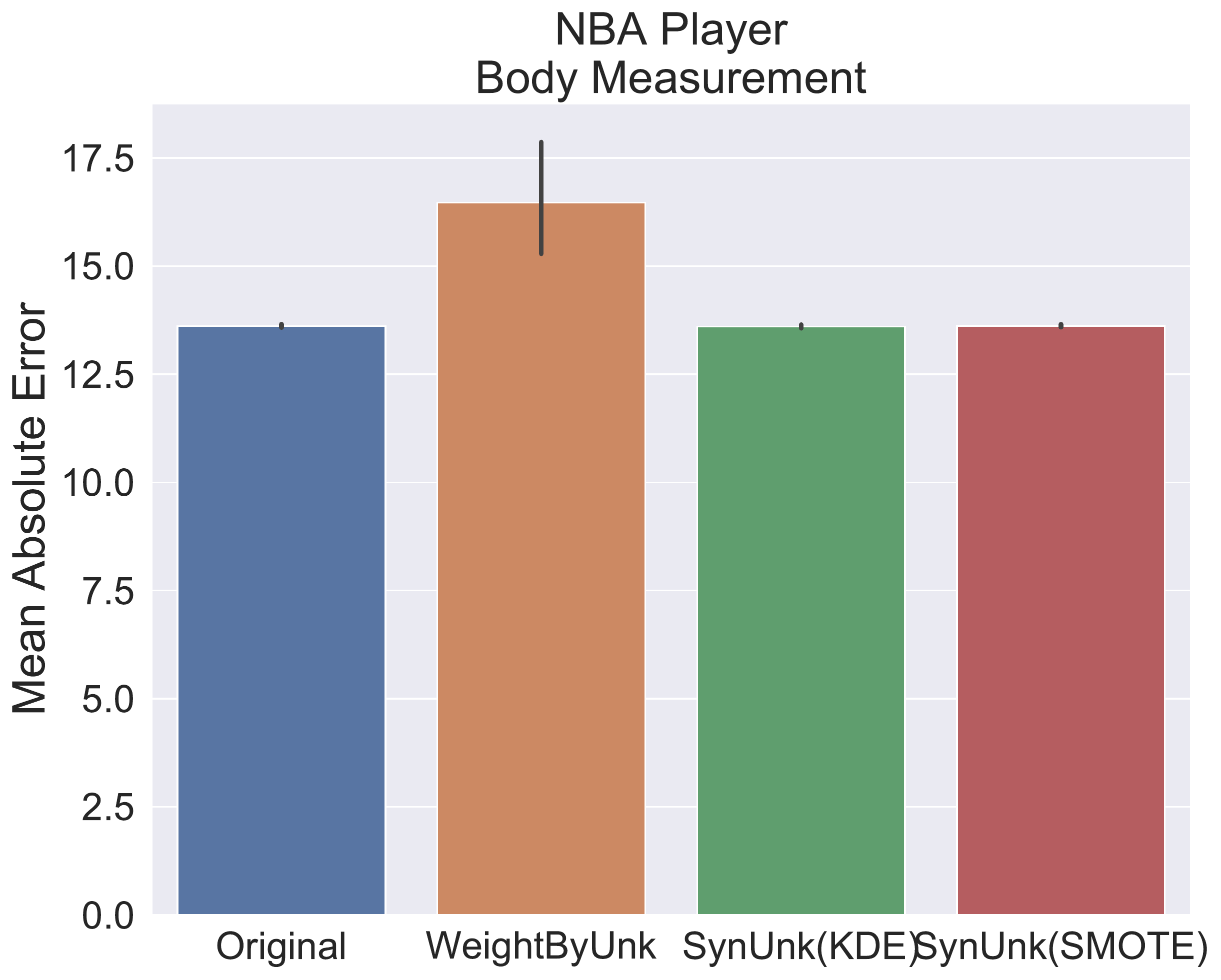}
\includegraphics[width=0.24\textwidth]{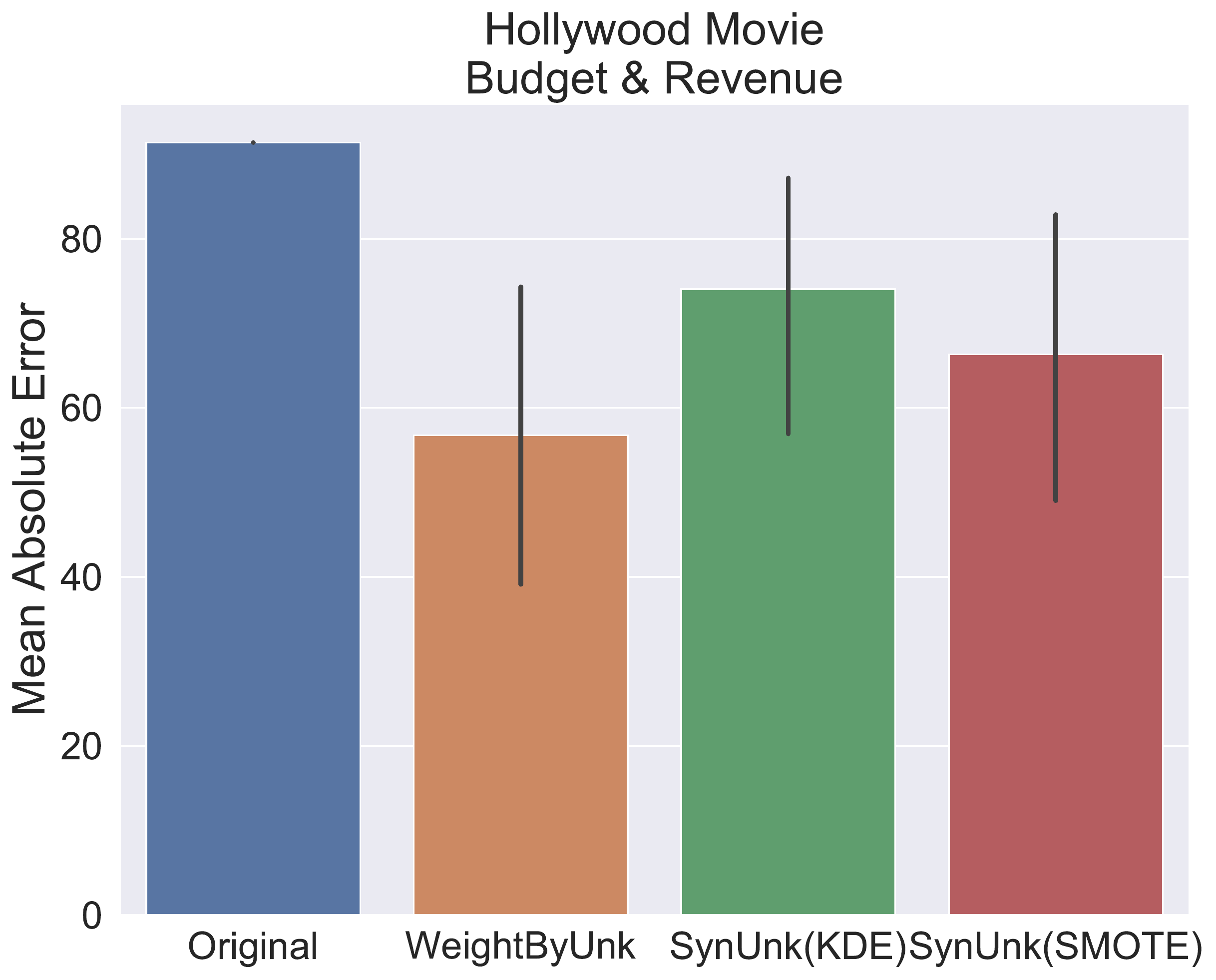}
 \caption{Mean absolute error on test datasets (the lower the better) for real-world crowdsourcing problems. Considering the unknown examples in training can improve the final model generalization (i.e., better test scores); there is no single unknown unknowns technique that works best in both cases.}
 \label{fig:unknownml:real_source1}
\end{figure}

\begin{figure*}[!t]
\centering
\includegraphics[width=\textwidth]{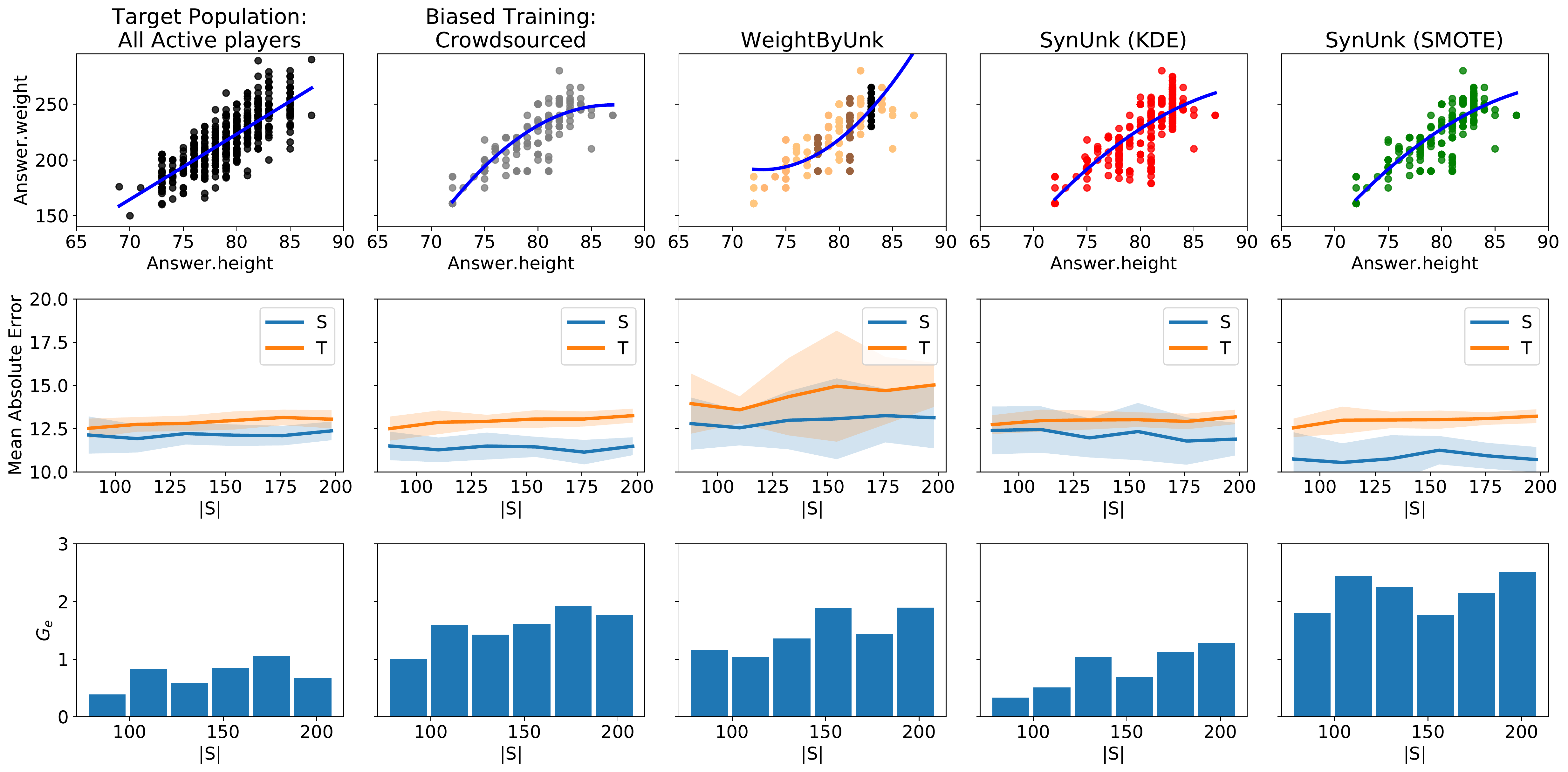}
 \caption{NBA players from 2015 to 2018 height vs. weight regression model evaluation. The training $S$ is crowdsourced and we use the ground truth dataset from NBA for testing $T$. The leftmost column (All Active NBA players from 2015 to 2018) is an ideal case where we train on $T$ ($S=T$); and we have a biased training case where the data is crowdsourced. We see that our technique, SynUnk (KDE), improves the model generalization error, $G_e$.}
 \label{fig:unknownml:real_source2}
\end{figure*}

For each example, we trained a simple degree-2 polynomial regression ML model. We first compare model accuracy, as measured by MAE, of the proposed approaches. The results from Figure~\ref{fig:unknownml:real_source1} show that our techniques can improve model accuracy relative to the baseline (Original).
As expected, we observe some degree of sampling bias in real-world data collection. To the extent that covariate shift or the sampling bias is systematic, we can potentially benefit from unknown unknowns. In particular, both SynUnk(KDE) and SynUnk(SMOTE) improve model accuracy relative to the baseline in the Movie example, and are roughly comparable to the baseline in the NBA example. Neither of these techniques performs significantly worse than the baseline, most likely because they are very conservative in estimating the unknown feature values.
It might seem that there is not much to be gained in the NBA example; however, our techniques can still help in terms of generalization error as shown in Figure~\ref{fig:unknownml:real_source2}.

Figure~\ref{fig:unknownml:real_source2} compares both the generalization accuracy and the generalization error $G_e$ for the various methods. We see that both SynUnk(KDE) and SynUnk(SMOTE) have comparable MAE to the biased Original dataset, so learning the unknown unknowns did not harm model performance. Moreover SynUnk(KDE), has the minimum generalization error $G_e$: results on test data are predictive of results on training data. 


\subsection{Which Technique To Use?}
Learning under covariate shift is not an easy problem, especially without any reference dataset (e.g., the unlabeled test data). 
Our experiments so far indicate that generating synthetic unknown examples will usually degrade a model's generalization ability by at most a small amount, and will often significantly improve generalization ability under a covariate shift. This is important because we want a technique that can be applied when the user does not know if there really is any covariate shift, since $p(x)$ and $T$ are unknown at training time. 

We expect the proposed unknown learning techniques to work well if a) unknown unknowns arise due to systematic biases in training data and b) the unknown unknowns have similar feature values to the rarely observed examples. If the unknown unknowns occur arbitrarily at random, and there exists no small bucket with high unknown unknowns concentration, then the unknown learning  effectively becomes a conservative oversampling. 
Moreover, if the proper model training requires observing extreme outliers or long tail of the missing distribution, it is not likely that the conservatively estimated unknown examples would help. Interestingly, these two conditions are also expected to be met for unknown unknowns discovery using an oracle \cite{lakkaraju2017identifying}.

\section{conclusion}
The ability to generalize beyond training data is critical for any practical learning algorithm. Assuming that the test data distribution will closely follow the training data distribution, many ML algorithms simply optimize for the best training score.
Unfortunately, this assumption is often risky under covariate shift, when  training and testing data are not sampled from the same distribution. 

In this work, we have developed novel techniques for learning the unknown examples that account for covariate shift. 
The key challenge is that we do not have an access to either test data or the target distribution at model training time.

Prior work for learning under covariate shift compared the training and the (unlabeled) testing data to detect and correct the shift. Instead, we use the fact that training data is often created by combining multiple sources with duplicate data; we apply species estimation techniques and explicitly model the missing unknown examples without using the test data. Our experimental results using simulations and real-world data indicate that the proposed techniques will typically not significantly hurt model generalization performance, and can dramatically help improve performance in the presence of covariate shift.

There are a number of interesting directions for future work. 
So far, we have focused on approaches that work directly with the data and treat the ML model as a black box. It would be interesting to try and understand how properties of specific ML algorithms relate to the unknown unknowns issue, and to exploit these properties to further improve generalization performance.

\bibliographystyle{IEEEtran}
\bibliography{main}


\end{document}